\documentclass[runningheads]{llncs}
\usepackage[T1]{fontenc}
\usepackage{booktabs}
\usepackage[misc]{ifsym}

\usepackage{mwe}

\usepackage[utf8]{inputenc}
\usepackage{graphicx}
\usepackage{caption}
\usepackage{subcaption}
\usepackage{booktabs}
\usepackage{amsmath}
\usepackage{dsfont}
\usepackage{algorithm}
\usepackage{algorithmic}
\usepackage{cite}
\usepackage{threeparttable}
\usepackage{url}
\usepackage{hyperref}
\usepackage{multirow}
\usepackage{pdfpages}
\usepackage{color}

\title{
Smart charging of large fleets of Electric Vehicles:  Independent Multi-Agent Reinforcement Learning approaches}
\titlerunning{EV Smart Charging:  Independent MARL approaches}
\author {Xavier Rate \inst{1} \and Eloann Le Guern \inst{2} \and Rapha\"el F\'eraud \inst{1}
\and Fatma Salem \inst{1}
\and Melissa Chiknoun \inst{1}
\and Eymeric Giabicani \inst{2}
\and Mehdi Feki \inst{1}
\and Patrick Maillé \inst{3}
\and Guy Camilleri \inst{4}
\and Anne Blavette \inst{2}
\and Hamid Benhamed \inst{2}
}

\institute{
Orange Research \email{@orange.com}
\and ENS Rennes \email{@ens-rennes.fr}
\and IMT Atlantique \email{@imt-atlantique.fr}
\and IRIT \email{@irit.fr}
}
\date{}

\begin{document}
\maketitle

\begin{abstract}
The electrification of transportation through electric vehicles introduces new challenges for power grid management, such as increased peak demand, voltage fluctuations, line overloads, and the integration of variable renewable energy sources. To enable efficient integration of EVs while minimizing costs for users and avoiding network overloads, implicit coordination between EVs is required. 
This work compares two independent multi-agent reinforcement learning approaches for optimizing such decentralized EV charging: contextual combinatorial bandits and policy gradient algorithms. Using a realistic simulation environment with autonomous agents making decisions based on local environmental information (including price signals, state-of-charge, and temporal constraints), we evaluate their performance across varying congestion levels, and mixed-strategy configurations with heterogeneous agent groups under dynamic electricity pricing derived from real photovoltaic production data.
\end{abstract}

\section{Introduction}

The widespread adoption of Electric Vehicles (EVs) poses significant challenges for modern electrical grids. With 17 million units sold in 2024, representing over 20\% of the market share, EVs are expected to continue their growth and account for 40\% of automotive sales by 2030~\cite{iea2025}. This rapid expansion has consequences for electrical networks because the global adoption of EVs comes with a rising electricity demand. In 2024, the global fleet of EVs consumed around 180 TWh of electricity, equivalent to Argentina's annual electricity consumption~\cite{iea2025}. 

While this growth is essential for global commitments to achieve carbon neutrality by 2050, it also raises concerns about grid stability. Uncoordinated charging during peak hours can cause voltage drops, line overloads, and network instability~\cite{tirunagari2022}.

Moreover, renewable energy sources often experience curtailment, with several GWh of clean electricity lost due to insufficient demand during peak production periods and the lack of energy storage solutions. For example, California's Independent System Operator regularly curtails renewable generation when supply exceeds demand, with solar curtailment reaching over 900 GWh in peak months during spring and fall~\cite{caiso2025}. In extreme cases, energy producers are even paid to stop generating electricity to prevent grid overload~\cite{bbc2024}. EV batteries can be leveraged as flexible storage to absorb excess energy and mitigate such curtailment issues.
Smart charging is therefore a promising solution, as it can reduce grid pressure while maximizing renewable energy utilization by utilizing the storage capacity of EV batteries. 
However, optimizing charging schedules for EV fleets requires handling two major issues: 
\begin {itemize}
\item scalability: the number of EVs is very large, which requires scalable algorithms,
\item uncertainty: the production of renewable energy and the consumption of energy depend on the weather, while
the availability of EVs depends on unknown human constraints.  
\end {itemize}

While centralized optimization, such as mixed-integer linear programming \cite {Franco2015},  requires full information about all EVs (arrival times, departure times, state of charges), prices, grid load, which is often impractical due to temporal, scalability, privacy, and communication constraints, fully decentralized approaches can scale with any number of EVs by allowing each EV to take decisions based on local and partial information. That is why, in this paper we will focus on fully decentralized approaches.

Reinforcement learning (RL) \cite {sutton2018} can optimize the scheduling of charging EVs, while handling uncertainties arising from EV owners' behavior, variability in renewable production and energy consumption.
Multi-Agent Reinforcement Learning (MARL) has become a popular approach for EV charging coordination~\cite{xie2025}. 
MARL algorithms based on 
the concept of centralized training and decentralized exploration \cite {rashid2018,lowe2017} are efficient in collaborative setting. However, the centralized critic needs the observations and the actions of all agents, which limits their scalability when the number of agents becomes large.
Independent Multi-Agent Reinforcement Learning \cite {tan1993} is a fully decentralized approach that can scale with any number of EVs, and is well suited for competitive setting: each agent has its own policy.
Proximal Policy Optimization (PPO) \cite {schulman2017proximalpolicyoptimizationalgorithms}, which is an improvement over Advantage Actor Critic (A2C) \cite {a2C}, is one of the most popular reinforcement algorithm due to its excellent compromise between performance and simplicity, making it a good candidate for each independent agent, while Simple Policy Optimization (SPO) \cite {SPO} is a recent version of PPO equipped with theoretical guarantees.
Based on a simplified version of reinforcement learning problem, Multi-armed bandits \cite {lattimore2020} benefit from a lighter and faster training that is a key advantage for real-world use in a changing environment.
Multi-Agent Multi-armed bandits have been successfully 
tested for optimizing communications in IoT networks \cite {Bonnefoi2018,Kerkouche2018,DAKDOUK2023}, and then 
for EV charging \cite {zafar2026,leguerndallo:hal-05262742}.
However, previous studies focus on just one approach without comparing different
algorithms. Moreover, real-world deployments will inevitably involve heterogeneous vehicle populations using different charging algorithms. As in 
Independent MARL, other agents are seen as a part of the environment, this heterogeneity between charging algorithms can change their performances.

In this paper, for bridging the gap, we investigate and compare the performances of Independent MARL based on modern policy gradient algorithms such as PPO \cite {schulman2017proximalpolicyoptimizationalgorithms}, A2C \cite {a2C}, and SPO \cite {SPO} with Linear Thompson Sampling \cite {DBLP:journals/corr/abs-1209-3352}
 for charging large fleets of vehicles.  
To assess algorithm robustness in real-world environment, we also evaluate mixed deployments of algorithms. 



\section{Problem Setting}

An electrical grid is a large graph that connects power plants to each customer. It involves different actors (producers, transporters, distributors, and consumers), each having different goals and constraints.
This study focuses on the decentralized smart charging of multiple electric vehicles connected to an electrical grid. We consider only one actor: the owners of electric vehicles, who seek to charge their vehicles at the lowest possible rate, when available. The electric grid constraints are taken into account through congestion.



\subsection{System Model}

\subsubsection{Electrical grid.}

We consider an elementary electrical grid that consists of $N$ electric vehicles connected to a transformer. 
The electrical grid is modeled without losses to focus on vehicle coordination, with total load being the sum of charging powers.
The transformer has the capacity to charge at the same time $K$ EVs, leading to congestion when demand exceeds this capacity ($N > K$).

\subsubsection{Electric Vehicle.}

We consider an episodic structure of time, where each day $d \in [D]$ is discretized into $T$ intervals. 
An EV $n \in [N]$ is available for charging during the time slots $\{a_d^n,...,b_d^n\} \in [T]^2$, where $a_d^n$ and $b_d^n$ are respectively the arrival and departure time slots, $a_d^n < b_d^n$. Arrival and departure time slots are dependent random variables.
Each EV $n \in [N]$ needs $m_d^n$ time slots to complete its charge during the day. $m_d^n$ is also a random variable.

\subsubsection{Pricing Model.}


We model the price at time slot $t$ as:
\begin {equation}
p(t) = 1 - \frac{P_{PV}(t)}{\bar{P}}
\label {price}
\end {equation}
where $P_{PV}(t)$ is the photovoltaic production at time slot $t$, normalized to the maximum of photovoltaic production $\bar{P}$. This pricing mechanism is designed to create incentives for renewable energy utilization: when solar production is at its peak ($P_{PV}(t) = \bar{P}$), the price is minimal ($p(t) = 0$), and when solar production is zero, the price reaches its maximum ($p(t) = 1$).
Notice, that the price depends on weather, and hence it is a random variable.


\subsection{Problem Formulation}

At the first glance, the decentralized smart charging problem can be viewed as a repeated planning problem over $D$ days, under uncertainty. 

Each day $d \in [D]$, each EV $n \in [N]$ aims to find a schedule of charging that allows to complete its charge at the end of the day, while minimizing its cost of charge.
Due to the availability of each EV $n$, and its initial state of charge, only some schedules $\mathbf{s}^n_d \in \{0,1\}^T$ are feasible:
\begin {equation}
\mathcal{S}_d^n = \left\{ \mathbf{s}^n_d \in \{0,1\}^T,  \text{supp}(\mathbf{s}^n_d) \subseteq \{a^n_d,...,b^n_d\}, \lVert \mathbf{s}^n_d \rVert_1 =m^n_d \right\},
\end {equation}
where $\text{supp}(\mathbf{s}^n_d) = \{t \in [T],\mathbf{s}^n_d(t)=1\}$, and $m_d^n$ denotes
the number of required charging time slots to reach the target state-of-charge.
Without taking into account the congestion due to other charging EVs, the optimal schedule for day $d \in [D]$, and agent $n \in [N]$ is:

\begin {equation}
\mathbf{s}^{n,*}_d = \arg\min_{s^n_d \in \mathcal{S}_d^n} {(\mathbf{s}^n_d})^ \top \mathbf{p}_d,
\end {equation}
where $\mathbf{p}_d \in [0,1]^T$ is the vector of price for day $d$.
The availability, the required number of time slots for charging, and the prices for the day are random variables. Moreover, the schedules of other EVs $i\neq n$ are unknown to EV $n$. If more than $K$ EVs choose the same time slot $t$ for charging, a congestion occurs $\mathbf{c}_d(t)=1$ and no EV can charge \footnote{congestion management can involve random disconnection of excess vehicles when the aggregate demand surpasses the transformer capacity, or congestion management can be done by available energy sharing.},  $\mathbf{c}_d(t) \in \{0,1\}^T$, else no congestion occurs $\mathbf{c}_d(t)=0$. 
We consider the case where all the EVs use a stochastic policy for choosing their charging time slots, and hence $\mathbf{c}_d(t)$ is a random variable.

This random variable $\mathbf{c}_d(t)$ drifts the decentralized charging problem from planning under uncertainty or model-based reinforcement learning to model free reinforcement learning. Indeed, choosing a good planning at the beginning of the day is not sufficient. When congestion occurs, for reaching their target state-of-charge, EVs need to replan their charging during the day. Moreover, the model free reinforcement learning framework allows to take into account imperfect observation of the state $\mathbf{x}^n_{d,t} \in \mathds {R}^l$, $l \in \mathds{N}^+$, through the policy that chooses charging of EV $n$ at time slot $t$.

At each time slot $t \in [T]$ of episode $d \in [D]$, each EV $n \in [N]$ chooses an action $\mathbf{s}^n_d(t) \in \{0,1\}$ until the end of the episode or until it is charged, i.e. $ ({\mathbf{s}^n_{d,. <t}})^\top (\mathbf{1-c}_{d,. <t})  = m_d^n$, where  $\mathbf{c}_d$ is the vector of $T$ dimensions where each coordinate is the event congestion, and the subscript $\mathbf{s}_{d,.<t}$ means the first $t$ dimensions of the vector $\mathbf{s}_d$.

The vector of reward $\mathbf{r}^n_d \in [0,1]^T$ of EV $n$ is provided at the end of the episode $d$: 
\begin {equation}
\mathbf{r}^n_d=(\mathbf{s}^n_d \circ (\mathbf{1}-\mathbf{c}_d))^\top (\mathbf{1}-\mathbf{p}_d), 
\end {equation}
where $\circ$ denotes the Hadamard product. The goal of each EV is to maximize its expected daily gain:
\begin {equation}
G^n(D)=\frac{1}{D}\sum_{d=1}^D \sum_{t=1}^T \mathds{E}[\mathbf{r}^n_d(t)],
\label {eq:gain}
\end {equation}
while minimizing the number of times the state of charge its not reached:
\begin {equation}
\sum_{d=1}^D \mathds{1}\{({\mathbf{s}^n_{d,. <T}})^\top (\mathbf{1-c}_{d,. <T})  < m_d^n\}.
\label {eq:SoCfail}
\end {equation}

%

\section {Independent Multi-Agent Reinforcement Learning approaches}

In this section, we present and adapt to our problem setting two reinforcement learning approaches that have been proposed in the literature for EV charging: Contextual Combinatorial Thompson Sampling \cite {zafar2026,leguerndallo:hal-05262742}, and policy gradient algorithms: PPO \cite {schulman2017proximalpolicyoptimizationalgorithms}, A2C \cite {a2C}, and SPO \cite {SPO}. 

\subsection{Contextual Combinatorial Linear Bandits}

\begin{algorithm}
    \caption{Contextual Combinatorial Bandits based on Linear Thompson Sampling for EV $n$}
    \begin{algorithmic}[1]
    \label{alg:context_combi_linTS}
    \renewcommand{\algorithmicrequire}{\textbf{Input:}}
    \renewcommand{\algorithmicensure}{\textbf{Output:}}
    \REQUIRE $\sigma > 0$, for all $t \in [T], A_t:=I_{l,l}, \mathbf{b}_t := 0_{l}, \hat{\boldsymbol{\theta}}_t := 0_{l}$
    \FOR {$d = 1,...,D$}
        \STATE  $m:=\min(m_d^n,b^n_d-a^n_d)$
        
        \FOR {$t=1,...,T$}
            \STATE Observe context $\mathbf{x}^n_{d,t}$
            \\ \textcolor{blue}{// Sampling parameter for each charging time slot}
            \STATE Sample $\tilde{\boldsymbol{\theta}_t} \sim \mathcal{N}(\hat{\boldsymbol{\theta}}_t, \sigma^2 A^{-1}_t)$
            \\ \textcolor{blue}{// Compute the likelihood of reward for each time slot}
            \STATE Compute $\tilde{\boldsymbol\mu}^n_{d}(t):=(\mathbf{x}^n_{d,t})^\top\boldsymbol{\tilde\theta}_t$
        \ENDFOR
        \STATE $t:=a^n_d$, $\mathbf{s}_d^n :=0_T$, $\mathbf{c}_d:=0_T$
        \WHILE {($t \leq b^n_d \text{ } \AND \text{ } m > 0$)}

            \STATE Define set $\mathcal{S}(t) := \{0,1\}^{b^n_d-t+1}$
            \\ \textcolor{blue}{// Choice of charging schedule}
            \STATE Choose $\mathbf{s}_t := \arg\max_{\mathbf{s} \in \mathcal{S}(t),\lvert\lvert \mathbf{s} \lvert\rvert_1 =m}\mathbf{s}^\top \tilde{\boldsymbol\mu}^n_{d,a^n_d \leq . < b^n_d-t+1}$
            \STATE Play action $\mathbf{s}_t(t)$, observe congestion $\mathbf{c}_d(t)$
            \STATE $\mathbf{s}_d^n(t):=\mathbf{s}_d^n(t)+\mathbf{s}_t(t)$
            \IF {$\mathbf{s}_t(t)=1 \text{ } \AND \text{ } \mathbf{c}_d(t) = 0 $}
                \STATE m:=m-1
            \ENDIF
            \STATE $t:=t+1$
        \ENDWHILE
    \STATE Observe price $\mathbf{p}_d$
    \\ \textcolor{blue}{// The linear models are updated for selected slots $\mathbf{s}_d^n(t)$}
    \FOR {$t=1,...,T$}
        \IF {$\mathbf{s}_d^n(t) = 1$}
            \STATE $\mathbf{r}^n_d(t):=\mathbf{s}_d^n(t) (1-\mathbf{c}_d(t)) (\mathbf{1}-\mathbf{p}_d(t))$
            \STATE $A_t := A_t + (\mathbf{x}^n_{d,t})^\top\mathbf{x}^n_{d,t}$
            \STATE $\textbf{b}_t:=\mathbf{b}_t+\mathbf{r}^n_d(t)\mathbf{x}^n_{d,t}$
            \STATE $\hat{\boldsymbol{\theta}}_t:= \textbf{b}_t^\top A_t^{-1}$
        \ENDIF
    \ENDFOR
    \ENDFOR
    \end{algorithmic} 
\end{algorithm}
A Multi-Armed Bandit (MAB) problem involves repeatedly selecting among actions, receiving rewards drawn from unknown probability distributions corresponding to the chosen actions~\cite{lattimore2020}. The goal is to maximize expected cumulative reward over time. 

As at the beginning of the game the player does not know which is the action with the highest mean reward, the player faces exploration-exploitation dilemma: in one hand, it has to explore actions with loosely estimated mean reward to built better estimates, and on the other hand, it has to exploit actions with highest estimated mean reward to gather rewards.

The decentralized charging problem can be viewed as a special case of the combinatorial bandit problem \cite {chen13}. Each day $d$, each EV $n$ has to choose $m^n_d$ time slots from $T$ time slots. When a congestion occurs, EV $n$ has to reselect the remaining time slots to reach the target state-of-charge. Moreover, at the beginning of each day $d$, EV $n$ can observe a context $\mathbf{x}_{d,t}^{n} \in \mathds {R}^l$ for each time slot $t$,  $l \in \mathds{N}^+$.

Algorithm \ref {alg:context_combi_linTS} uses a Linear Thompson Sampling (LinTS) \cite{DBLP:journals/corr/abs-1209-3352} for taking into account the context 
to evaluate the highest likelihood of reward at time slot $t$ (line 6).
Then, the schedule of charging is selected using the time slots with 
the highest likelihood of reward (line 11), where
$\tilde{\boldsymbol\mu}^n_{d,a^n_d \leq . < b^n_d}$ means the dimensions between $a_d^n$ and $b_d^n$ of the vector
$\tilde{\boldsymbol\mu}^n_d$.
The computational complexity of the selection of the best schedule is linear in term of number of time slots $T$. Moreover, this selection could be done at the first time slot and then only if a congestion occurs at the previous time slot. The price of each time slot is revealed at the end of the day (line 19). It allows to update the linear model of each selected time slot (lines 22-25).

\subsection{Proximal Policy Optimization (PPO)}

In contrast to Contextual Combinatorial Linear Bandits, standard Reinforcement Learning addresses sequential decision-making problems modeled 
as Markov Decision Processes: an agent observes a state, select 
an action according to its policy, and receives a reward that depends on the current state, the played action, and the following state, while in Contextual Combinatorial Linear Bandits the reward depends only on the state and on the played action. 

\begin{algorithm}
\caption{PPO}
\begin{algorithmic}[1]
\label {alg:ppo}
\REQUIRE Initial policy $\pi_{\boldsymbol\theta_0}$, value function $V_{\boldsymbol\phi_0}$
\REQUIRE Hyperparameters: $\epsilon$ (clipping), $\lambda$ (GAE), $K$ (epochs), $D$ (horizon), $M$ (mini-batch size) 
\STATE $\mathcal{D}:=\emptyset$
\\ \textcolor{blue}{// Collect trajectories}
\FOR{iteration $d = 1, 2, \ldots, D$}    
    \STATE Collect trajectories $\mathcal{D} := \{ (\mathbf{x}^n_{d,t}, \mathbf{s}_d^n(t), \mathbf{c}_d(t), \mathbf{p}_{d}(t)) \}$ by running $\pi_{\theta_{old}}$   
    
    \textcolor{blue}{// Compute returns and advantages for each timestep}
    \FOR {each trajectory $\mathcal{T}$ in $\mathcal{D}$}
    \FOR{each $t \in \mathcal{T}$}
        \STATE $\mathbf{r}^n_d(t) := \mathbf{s}_d^n(t)(1-\mathbf{c}_d(t)(1-\mathbf{p}_{d}(t))$
        \STATE $R(t) := \sum_{l=0}^{|\mathcal{T}|-1-t}  \mathbf{r}^n_d(t+l)$
        \STATE $\delta(t) := \mathbf{r}^n_d(t) + V_\phi(\mathbf{x}^n_{d,t+1}) - V_\phi(\mathbf{x}^n_{d,t})$
        \STATE $\hat{A}(t) := \sum_{l=0}^{|\mathcal{T}|-1-t} (\lambda)^l \delta(t+l)$
    \ENDFOR
    \ENDFOR
    \\\textcolor{blue}{// Optimize over multiple epochs}
    \FOR{epoch $k = 1$ to $K$}
        \STATE Split $\mathcal{D}$ into $M$ mini-batches
        \\ color{blue}{// Compute loss averaged over timesteps in batch}
        \FOR{each mini-batch $\mathcal{B} = \{(\mathbf{x}^n_{d,t}, \mathbf{s}_d^n(t), \hat{A}(t), {R}(t))\}_{t \in \mathcal{B}}$}
            \STATE $\rho_{\boldsymbol\theta}(t) := \frac{\pi_{\boldsymbol\theta}(\mathbf{s}_d(t)|\mathbf{x}^n_{d,t})}{\pi_{\boldsymbol\theta_{old}}(\mathbf{s}_d(t)|\mathbf{x}^n_{d,t})}$
            \STATE $L^{CLIP} := \frac{1}{|\mathcal{B}|}\sum_{t \in \mathcal{B}} \min(\rho_{\boldsymbol\theta}(t)\hat{A}(t), \text{clip}(\rho_{\boldsymbol\theta}(t), 1-\epsilon, 1+\epsilon)\hat{A}(t))$
            
            \STATE $L^{VF} := \frac{1}{|\mathcal{B}|}\sum_{t \in \mathcal{B}} (V_\phi(\mathbf{x}^n_{d,t}) - {R}(t))^2$
            
            \STATE Update $\boldsymbol\theta$ and $\boldsymbol\phi$ using gradients of $L^{CLIP}$ and $L^{VF}$
        \ENDFOR
    \ENDFOR
    \STATE $\mathcal{D}:=\emptyset$, ${\boldsymbol\theta}_{old} := \boldsymbol\theta$
\ENDFOR
\end{algorithmic}
\end{algorithm}

The decentralized charging problem can be viewed as a reinforcement learning problem, where the environment is composed by the considered EVs, and the electrical grid, where the terminal states are the maximum length $T$ of the episode and the event "target state-of-charge is reached", and where the discount factor is $1$: the reward of the episode is the sum of rewards of charging time slots.

Proximal Policy Optimization (PPO) \cite {schulman2017proximalpolicyoptimizationalgorithms} is a policy gradient algorithm that addresses the instability of traditional policy gradient methods by limiting  policy updates. 
Algorithm \ref {alg:ppo} collects trajectories using the previously updated policy $\pi_{old}$ (line 3), computes  advantages using Generalized Advantage Estimation (GAE) (line 6-9).
PPO uses collected trajectories for multiple optimization epochs with  mini-batch gradient descent (lines 14-19)
The key innovation is the clipping mechanism: the probability ratio between new and old policies $\rho_{\boldsymbol\theta}$ (line 15) is constrained within $[1-\epsilon, 1+\epsilon]$, ensuring conservative updates (line 16). 
PPO jointly trains an actor (policy) $\pi_{\boldsymbol\theta}$ and a critic (value function) $V_{\boldsymbol\Phi}$ (line 18).

\subsection {Advantage Actor Critic (A2C)}

PPO and A2C differ only by the clipping mechanism (lines 15-16 Algorithm \ref {alg:ppo}). This  gives PPO greater stability during gradient descent, thus reducing overfitting and allowing it to replay $K$ times the buffer of trajectories $\mathcal{D}$ (line 12), thereby improving sample efficiency.

\subsection {Simple Policy Optimization (SPO)}

While PPO uses hard clipping  (line 16 Algorithm \ref{alg:ppo}), resulting in zero gradients for data points outside the interval $[1-\epsilon,1+\epsilon]$, SPO \cite {SPO} replaces this with soft quadratic regularization, which maintains non-zero gradients everywhere. In practice, the only difference with the PPO algorithm lies in line 16:

$L_p := -\frac{1}{|\mathcal{B}|} \sum_{t \in \mathcal{B}} \left\{ \rho_\theta(t) \hat{A}(t) - \frac{|\hat{A}(t)|}{2\epsilon} \cdot \left[\rho_\theta(t) - 1\right]^2 \right\}
$

\section {Experimental setup}

\subsection{Simulation Environment}

We developed a multi-agent EV charging simulation environment in Python \footnote{a gitlab can be found here: https://gitlab.com/lpplc/anon}, leveraging our custom multi-agent platform built on \textit{Gymnasium}~\cite{towers2024gymnasiumstandardinterfacereinforcement}, building upon the work of~\cite{leguerndallo:hal-05262742}.
Key features include:
\begin{itemize}
    \item configurable capacity $K$ of electrical grid,
    \item support for heterogeneous strategies across the vehicle population,
    \item integration of historical photovoltaic production data,
    \item customizable distributions for EV arrivals and departures,
    \item congestion management by random disconnection of excess vehicles when
the aggregate demand surpasses the transformer capacity,
\end{itemize}

\subsection{Electric Vehicle Parameters}

Each vehicle has a $22$~kW charging power and a $92$~kWh battery capacity~\cite{statista2025}. Vehicle arrivals follow a truncated normal distribution centered at 9{:}00~AM ($\tau_i^{\text{arr}} \sim \text{TruncNorm}(\mu=9, \sigma=1.5, a=6, b=12)$), while parking durations follow a truncated exponential distribution ($D_i \sim \text{TruncExp}(\lambda=1, a=7, b=10)$ hours). The initial state of charge varies uniformly between 10\% and 70\%, with a target SoC of 80\%.

\subsection{Data and Electricity Prices}

Four years (2021-2024) of photovoltaic production data available on Elia's portal~\cite{elia2025data} are used to derive electricity price using equation (\ref {price}).
Two information are considered: one with perfect real-time solar data and another with realistic day-ahead forecasts including uncertainty. The data are then normalized between the 1st and 99th percentiles with a 15-minute granularity.
$4000$ days are generated by sampling uniformly these four years of electricity price. $3000$ days are used for training for training dataset and $1000$ days are used for test dataset.

\subsection{Algorithms and baselines Implementation}

\subsubsection{Greedy.}

This strategy sorts all available time slots between arrival and departure times in ascending order of electricity price, then selects the cheapest time slots to meet the charging demand. If the current time falls within this selection, the vehicle charges; otherwise, it waits. 
Without congestion and the true electricity price, this policy is optimal.
If a charging attempt fails due to congestion, the selection process is repeated to find the most suitable time slots. The electricity price may be based on a forecast or the actual price.

\subsubsection{Bernoulli Thompson Sampling (TS).}

Bernoulli Thompson Sampling is an optimal algorithm for the stochastic Multi-Armed Bandits \cite {kaufmann2012}. It does not take into account the context $\mathbf{x}^n_{d,t} \in \mathds{R}^l$. It is used as a baseline to evaluate the gain of contextual information in Linear Thompson Sampling.
At each timestep, Bernoulli Thompson Sampling models the probability of successfully charging (i.e., not being congested) as a Beta distribution:
\[
p_{\text{success}}(t) \sim \text{Beta}(\alpha, \beta).
\]

Parameters are updated according to observed outcomes:
\[
\alpha:= \alpha + 1-\mathbf{c}_d(t),
\qquad
\beta[t] \leftarrow \beta[t] + \mathbf{c}_d(t).
\]

At each time slot, the algorithm samples from these Beta distributions to estimate the probability of successful charging and computes an expected gain for each time slot as:
\[
\hat{\mathbf{g}}^n_d(t) = p_{\text{success}}(t) (1 - \hat{\mathbf{p}}_d(t)),
\]
where $\hat{\mathbf{p}}_d(t)$ is forecast price at time slot $t$. The algorithm then selects $m^n_d$ time slots with the highest expected gain.
If congestion occurs, the selection process is repeated. The parameters $\alpha$ and $\beta$ are initialized to $1$.

\subsubsection{Linear Thompson Sampling (LinTS).}

Linear Thompson Sampling  \cite{DBLP:journals/corr/abs-1209-3352} is used in Algorithm \ref {alg:context_combi_linTS} for selecting the best schedules. The variance parameter $v$ is set to $1$.
The context $\mathbf{x}^n_{d,t}$ contains the forecast price, the current time step, the current week. Given the cyclic nature of time step and week a $sin/cos$ coding is used.

\subsubsection{Neural Thompson Sampling (NeuralTS).}

This algorithm is the same that Algorithm \ref {alg:context_combi_linTS}, except that instead using LinTS it employs a neural network $f_{\boldsymbol{\theta}}$ to model the relationship between contextual features $\mathbf{x}^n_{d,t}$ and expected reward $\mathbf{r}^n_d(t)$. The neural network uses a single hidden layer with $64$ neurons, a learning rate of $10^{-3}$, and Sherman-Morrison updates for the Bayesian linear head. The context $\mathbf{x}^n_{d,t}$ is the same as the one used in LinTS.

\subsubsection{Proximal Policy Optimization (PPO).}

PPO~\cite{schulman2017proximalpolicyoptimizationalgorithms} is used in Algorithm \ref {alg:ppo} for selecting the best schedules.
PPO uses $25$ epochs per update, a batch size of $512$, a cliping range of $0.3$, and a learning rate of $5\times 10^{-4}$. PPO is implemented with Stable-Baselines3~\cite{stable-baselines3}.
The state $\mathbf{x}^n_{d,t}$ used in PPO contains the same information as that of LinTS, 
plus plug status, which is an indicator of connection to the electrical grid, the state of charge, and the remaining number of time steps. This additional contextual information is directly included in Contextual Combinatorial MAB (Algorithm \ref {alg:context_combi_linTS}).

\subsubsection{Advantage Actor Critic (A2C).} A2C \cite {a2C} uses the same parameters, contexts and implementation as PPO.

\subsubsection{Simple Policy Optimization (SPO).} SPO \cite {SPO} uses the same parameters, contexts and implementation as PPO.

\section{Experimental Results}
\label {sec:Results}

\subsection{Evaluation Protocol}

All algorithms are trained over $3000$ episodes, each representing a one-day scenario, and tested on $1000$ unseen episodes. 
Performance is assessed using two complementary metrics: the average daily rewards achieved by the agents (equation \ref {eq:gain}) and the number of "SOC failure" events (equation \ref {eq:SoCfail}). 
Each experiment is done $5$ times.

We consider two scenarios within this unified protocol. The first scenario corresponds to homogeneous deployments, where all 20 EVs use the same algorithm under different capacity levels and both information regimes (perfect real-time prices versus day-ahead forecasts). 
The second scenario involves heterogeneous deployments, where multiple algorithms coexist within the same environment to assess robustness and interoperability.

\subsection{Homogeneous Groups}

\subsubsection{Learning curves.}

\begin{figure}[h!]
\centering
\includegraphics[width=\textwidth]{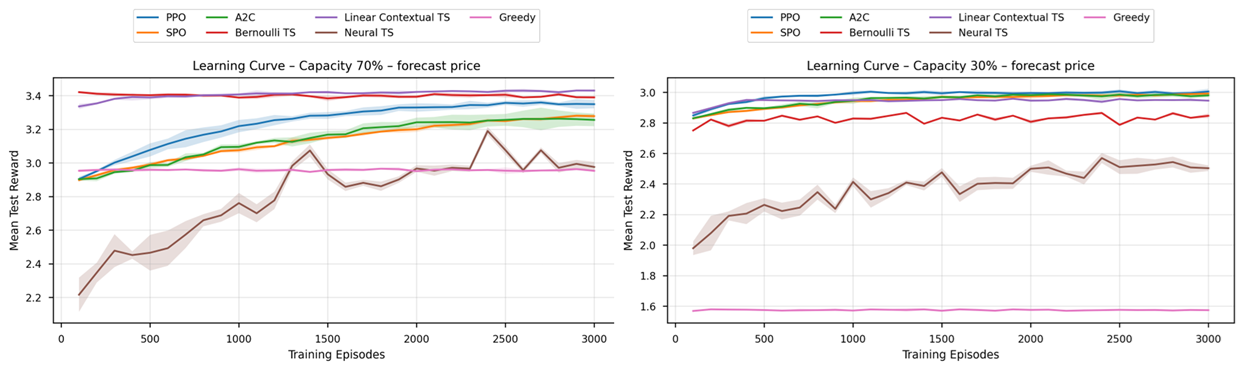}

\caption{Learning curves. Left panel $70 \%$ capacity. Right panel $30 \%$ capacity.}
\label{fig:learning_curves}
\end {figure}

We focus first on the learning curves to study the convergence time of different learning algorithms, which is crucial in a constantly evolving real-world environment.

In Figure \ref {fig:learning_curves}, for each algorithm, the mean reward evaluated on the test dataset versus the number of training episodes is plotted
for a grid capacity allowing simultaneous charging $70 \%$ of EVs (left panel) and $30 \%$ of EVs (right panel). 
In low congestion regime (capacity of $70 \%$), EVs equipped with LinTS and and TS converge to a reasonably good policy in few episodes. Then, due to additional contextual information, EVs equipped with LinTS outperforms those with TS in term of mean reward.
EVs equipped with PPO, SPO and A2C do not have time to converge since their test mean reward still increase after $3000$ episodes. Neural Thompson Sampling exhibits a very slow convergence.
In high congestion regime (capacity of $30 \%$), EVs equipped with LinTS also quickly converge to a reasonably good policy that is improved after $400$ episodes. 
After $500$ episodes, EVs equipped with PPO, SPO or A2C exhibit slightly better performance than those using LinTS. EVS equipped with TS converge very fast, but are outperformed by policies that use contextual information. NeuralTS would require more episodes to converge.

\subsubsection{Performance Analysis.}

\begin{figure}[h!]
\centering
\begin{subfigure}[b]{\textwidth}
    \centering
    \includegraphics[width=\textwidth]{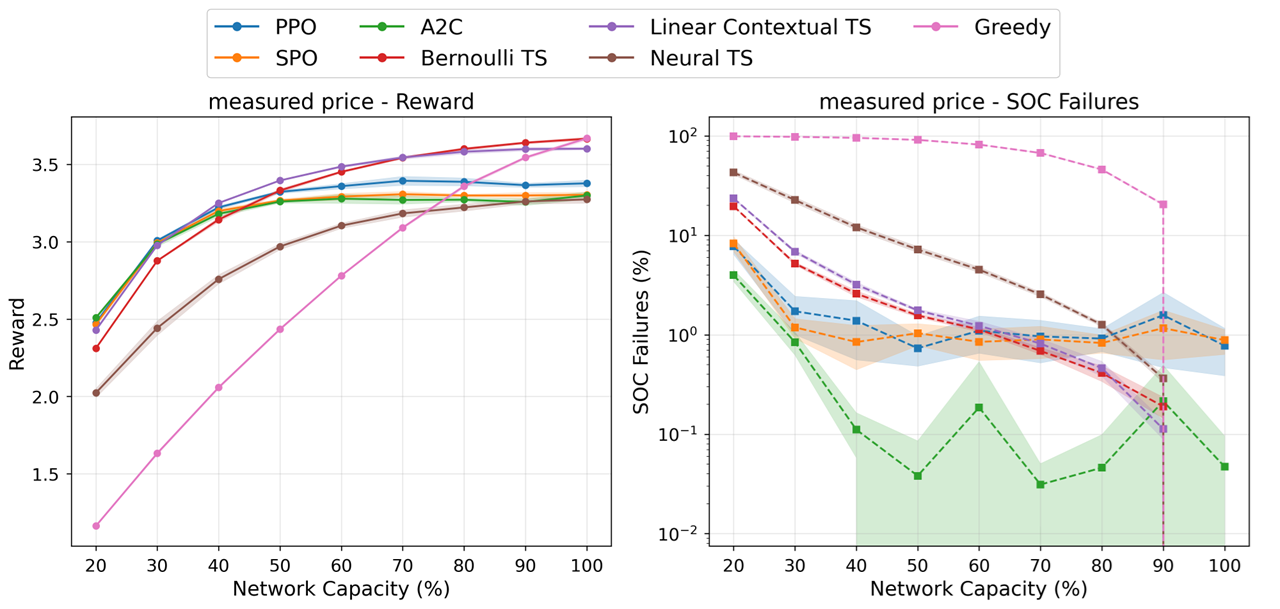}
    \caption{Measured prices}
    \label{fig:performance_measured}
\end{subfigure}
\vfill
\begin{subfigure}[b]{\textwidth}
    \centering
    \includegraphics[width=\textwidth]{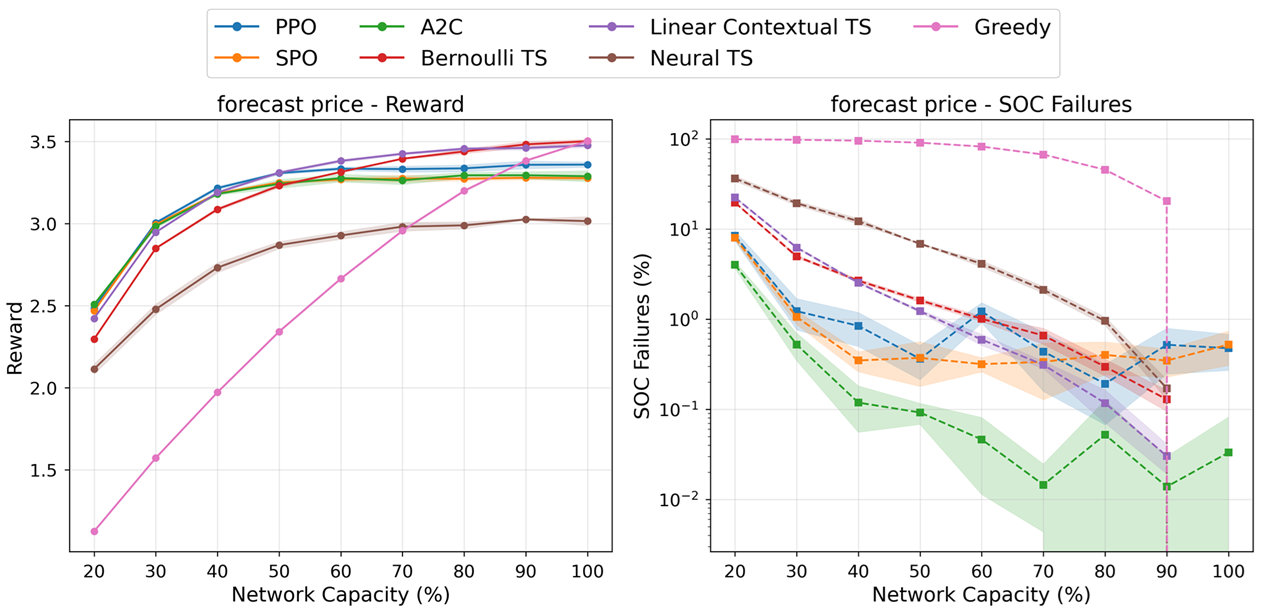}
    \caption{Day-ahead forecasted prices}
    \label{fig:performance_dayahead}
\end{subfigure}
\caption{Algorithm performance for homogeneous groups across capacity levels. Left panels: average rewards per vehicle; right panels: SOC failure rates (log scale).}
\label{fig:performance_homogeneous}
\end{figure}

To analyze the performance of algorithms, in Figure \ref {fig:performance_homogeneous}, we plot the average reward and the percentage of failures in the state-of-charge measurement against capacity levels (percentage of electric vehicles that can be charged simultaneously).
Results are presented for both measured prices in Figure \ref {fig:performance_measured} (perfect real-time information) and day-ahead forecasted prices in Figure \ref {fig:performance_dayahead} (realistic uncertainty).

While the performance of greedy baseline is optimal when the price information is perfect and when there is no congestion (capacity level $100 \%$), we observe that its performance in terms of mean reward and SOC failures are the worst for low capacity level (Figure \ref {fig:performance_measured}).
Notice that in perfect condition (capacity level $100 \%$ and measured price), TS and LinTS obtain near optimal performance, while PPO, SPO and A2C do not achieve this performance. This is consistent with their lower convergence rate observed in Figure \ref {fig:learning_curves}a.
For low capacity (between $20 \%$ and $30 \%$), EVs equipped with PPO, SPO or A2C slightly outperform those equipped with LinTS, while TS that does not use contextual information and NeuralTS that does not finish its convergence are outperformed.
In terms of SOC failures, EVs equipped with A2C, PPO and SPO outperform those equipped with LinTS for low to medium capacity (between $20 \%$ and $50 \%$). 
This suggests that EV equipped with policy gradient algorithms coordinate better than other algorithms. For high capacity (between $70 \%$ and $100 \%$), EVs equipped with LinTS and A2C dominate other algorithms.
The worst algorithm is Greedy that chooses the lower prices and hence, as soon as there is congestion, fails to charge EVs.
Finally, while the performances of LinTS and TS decrease when the electricity prices are forecasted, those of PPO, SPO, and A2C are the same. 

Globally, LinTS exhibits a fast convergence (Figure \ref {fig:learning_curves}), and good performances in terms of electricity price and SOC failures for all network capacity levels (Figure \ref {fig:performance_homogeneous}), while policy gradient algorithms are better for low capacity, which suggests that EV equipped with PPO, SPO, or A2C coordinate better than other algorithms.

\subsection{Heterogeneous Groups}

Real-world deployments will inevitably involve heterogeneous vehicle populations using different charging algorithms. To assess algorithm robustness in such environments, we evaluate mixed deployments with $10$ PPO agents and $10$ LinTS agents sharing the same network constraints.

\begin{figure}[H]
\centering
    \includegraphics[width=\textwidth]{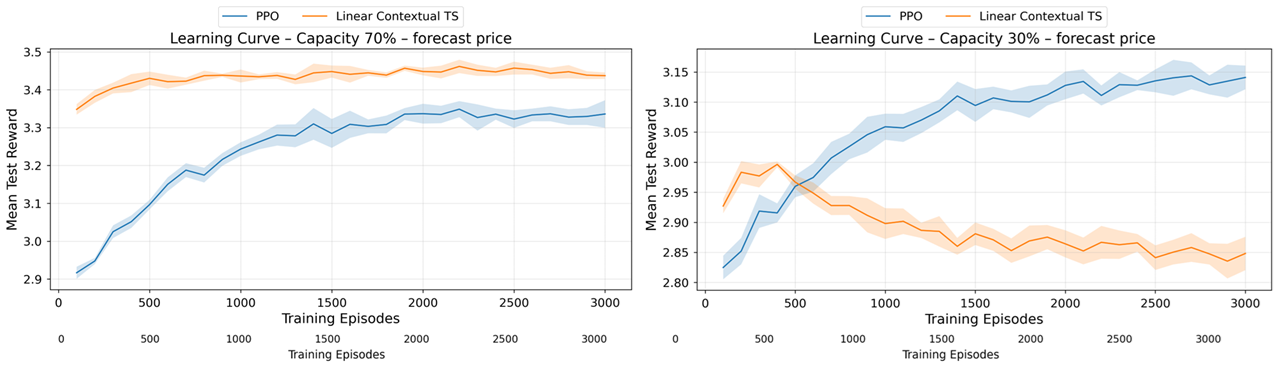}
\caption{Learning curves for heterogeneous groups ($10$ PPO agents, $10$ LinTS agents). Left panel $70 \%$ capacity. Right panel $30 \%$ capacity.}
\label{fig:learning_curves_hetero}
\end{figure}

\begin{figure}[H]
\centering
    \includegraphics[width=\textwidth]{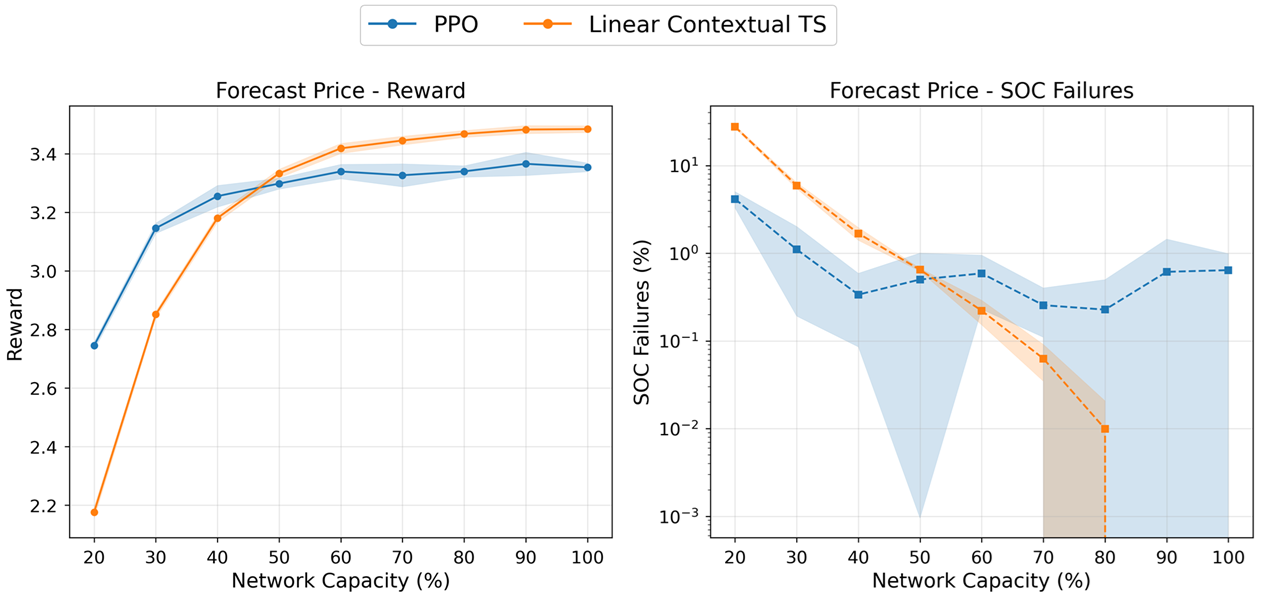}
\caption{Left panel: average rewards per vehicle; right panel: SOC failure rates (log scale). Heterogeneous groups ($10$ PPO agents, $10$ LinTS agents) across capacity levels.}
\label{fig:performance_hetero}
\end{figure}

For high network capacity ($70 \%$), the conclusion of the previous experiment with homogeneous groups of EVs is reinforced (Figure \ref {fig:learning_curves}): EVs equipped with LinTS converge faster to a good solution that is slightly improved during time (figure \ref {fig:learning_curves_hetero} left panel).
However, for low network capacity ($30 \%$) the behavior of the learning curve of EVs equipped with LinTS radically change: after quickly reaching a good solution, the mean reward on the test set decreases (Figure \ref {fig:learning_curves_hetero} right panel).
This is due to the learning of electric vehicles equipped with PPO, which, when LinTS-equipped electric vehicles focus on certain time slots, manage to find a better compromise between price and congestion, thus polluting the solution obtained by LinTS.
Indeed, LinTS is based on a stationary stochastic assumption that does not hold in high congestion regime in face of another group of learning agents, while standard reinforcement learning handles dynamic environment. 
It should be noted, however, that EVs equipped with PPO only demonstrate superior performance to those equipped with LinTS after $500$ days, or nearly a year and a half. Furthermore, EVs equipped with LinTS  outperform those equipped with PPO on medium to high capacity grids (Figure \ref {fig:performance_hetero}).

\section {Conclusion}

To develop a scalable approach for optimizing electric vehicle charging, we proposed two approaches based on independent multi-agent reinforcement learning: contextual and combinatorial linear bandits, and classical reinforcement learning algorithms such as PPO, SPO, and A2C. When the agents use the same learning algorithm, Contextual and combinatorial linear bandits based on LinTS  find a good solution faster, which, in the absence of congestion, achieves performance close to the one of the optimal policy, and, in the presence of significant congestion, closely matches those of the best classical reinforcement learning algorithms.
However, when two teams of agents compete to recharge their batteries on the same highly congested electrical grid, we have observed that strong learners such as PPO, SPO or A2C can significantly outperform LinTS. 
However, the convergence speed of these algorithms remains far too slow to be learned online in even a slightly changing environment (one year and a half to do better than LinTS). 
Using a sliding window for LinTS might seem like a good solution for improving performance against an opposing team. Indeed, the sliding window allows players to forget part of their past performance and thus adapt to a change \cite {Garivier2011}. However, electricity consumption and photovoltaic electricity production are seasonally dependent. This means that to learn this dependence, a sliding window of at least one year is necessary, which severely limits the ability to adapt to a change in the opposing team's strategy. Future works will focus on adversarial linear bandits \cite {neu2020}, which should allow for adaptation to changes and could also provide theoretical guarantees.

\bibliographystyle{ieeetr}
\bibliography{references}

@InProceedings{Garivier2011,
author="Garivier, Aur{\'e}lien
and Moulines, Eric",
title="On Upper-Confidence Bound Policies for Switching Bandit Problems",
booktitle="Algorithmic Learning Theory",
year="2011"
}

@InProceedings{neu2020,
  title = 	 {Efficient and robust algorithms for adversarial linear contextual bandits},
  author =       {Neu, Gergely and Olkhovskaya, Julia},
  booktitle = 	 {Proceedings of Thirty Third Conference on Learning Theory},
  year = 	 {2020}
}

@inproceedings{SPO,
  title={Simple Policy Optimization},
  author={Xie, Zhengpeng and Zhang, Qiang and Yang, Fan and Hutter, Marco and Xu, Renjing},
  booktitle={Proceedings of the 42nd International Conference on Machine Learning},
  year={2025}
}

@article{a2c,
  author       = {Volodymyr Mnih and
                  Adri{\`{a}} Puigdom{\`{e}}nech Badia and
                  Mehdi Mirza and
                  Alex Graves and
                  Timothy P. Lillicrap and
                  Tim Harley and
                  David Silver and
                  Koray Kavukcuoglu},
  title        = {Asynchronous Methods for Deep Reinforcement Learning},
  journal      = {arXiv},
  year         = {2016}
}

@InProceedings{Bonnefoi2018,
author="Bonnefoi, R{\'e}mi
and Besson, Lilian
and Moy, Christophe
and Kaufmann, Emilie
and Palicot, Jacques",
title="Multi-Armed Bandit Learning in IoT Networks: Learning Helps Even in Non-stationary Settings",
booktitle="Cognitive Radio Oriented Wireless Networks",
year="2018"
}

@INPROCEEDINGS{Kerkouche2018,
  author={Kerkouche, Raouf and Alami, Reda and F\'eraud, Raphaël and Varsier, Nadege and Maill\'e', Patrick},
  booktitle={2018 25th International Conference on Telecommunications (ICT)}, 
  title={Node-based optimization of LoRa transmissions with Multi-Armed Bandit algorithms}, 
  year={2018}
}

@inproceedings{kaufmann2012,
  title={Thompson Sampling: An Asymptotically Optimal Finite-Time Analysis},
  author={Kaufmann, Emilie and Korda, Nathaniel and Munos, R{\'e}mi},
  booktitle={International Conference on Algorithmic Learning Theory},
  year={2012}
}

@InProceedings{chen13,
  title = 	 {Combinatorial Multi-Armed Bandit: General Framework and Applications},
  author = 	 {Chen, Wei and Wang, Yajun and Yuan, Yang},
  booktitle = 	 {Proceedings of the 30th International Conference on Machine Learning},
  year = 	 {2013}
}

@inproceedings{tan1993,
  title     = {Multi-Agent Reinforcement Learning: Independent vs. Cooperative Agents},
  author    = {Tan, Ming},
  booktitle = {Proceedings of the Tenth International Conference on Machine Learning},
  year      = {1993}
}

@article{Franco2015,
  title   = {A Mixed-Integer Linear Programming Model for the Electric Vehicle Charging Coordination Problem in Unbalanced Electrical Distribution Systems},
  author  = {J. F. Franco, M. J. Rider,  R. Romero},
  journal = {IEEE Transactions on Smart Grid},
  year    = {2015}
}

@book{lattimore2020,
  title     = {Bandit Algorithms},
  author    = {Lattimore, Tor and Szepesv{\'a}ri, Csaba},
  year      = {2020},
  publisher = {Cambridge University Press},
  note      = {Available online at \url{https://banditalgs.com}}
}

@inproceedings{lowe2017,
  title     = {Multi-Agent Actor-Critic for Mixed Cooperative-Competitive Environments},
  author    = {Lowe, Ryan and Wu, Yi and Tamar, Aviv and Harb, Jean and Abbeel, Pieter and Mordatch, Igor},
  booktitle = {Advances in Neural Information Processing Systems},
  series    = {NeurIPS},
  volume    = {30},
  year      = {2017}
}

@inproceedings{rashid2018,
  title     = {QMIX: Monotonic Value Function Factorisation for Deep Multi-Agent Reinforcement Learning},
  author    = {Rashid, Tabish and Samvelyan, Mikayel and de Witt, Christian Schroeder
               and Farquhar, Gregory and Foerster, Jakob and Whiteson, Shimon},
  booktitle = {Proceedings of the 35th International Conference on Machine Learning},
  series    = {ICML},
  year      = {2018},
  volume    = {80},
  pages     = {4295--4304},
  publisher = {PMLR}
}

@article{iea2025,
  title={Global EV Outlook 2025},
  author={{IEA}},
  year={2025},
  publisher={IEA, Paris},
  url={https://www.iea.org/reports/global-ev-outlook-2025},
  note={Licence: CC BY 4.0}
}

@article{tirunagari2022,
  author={Tirunagari, Sridevi and Gu, Mingchen and Meegahapola, Lasantha},
  title={Reaping the Benefits of Smart Electric Vehicle Charging and Vehicle-to-Grid Technologies: Regulatory, Policy and Technical Aspects},
  journal={IEEE Access},
  year={2022},
  volume={10},
  pages={114657-114672},
  doi={10.1109/ACCESS.2022.3217525}
}

@book{sutton2018,
  author={R. Sutton and A. Barto},
  title={Reinforcement Learning: An Introduction},
  edition={2nd},
  publisher={MIT Press},
  year={2018}
}

@article{dakdouk2023,
  author={H. Dakdouk and R. Feraud and N. Varsier and P. Maill\'e and R. Laroche},
  title={Massive multi-player multi-armed bandits for IoT networks: An application on LoRa networks},
  journal={Ad Hoc Networks},
  year={2023}
}

@article{zafar2026,
  author={S. Zafar and R. F\'eraud and A. Blavette and G. Camilleri and H. Ben Ahmed},
  title={Decentralized multi-agent multi-armed bandits for smart electric vehicles charging},
  journal={Engineering Applications of Artificial Intelligence},
  year={2026}
}

@inproceedings{leguerndallo:hal-05262742,
  TITLE = {{Multi-Agent Multi-Armed Bandits: application to EVs smart charging with grid constraints}},
  AUTHOR = {Le Guern-Dall'o, Eloann and F{\'e}raud, Rapha{\"e}l and Camilleri, Guy and Maill{\'e}, Patrick and Petit, Fabien and Zorgati, Riadh and Ben Ahmed, H. and Blavette, Anne},
  URL = {https://hal.science/hal-05262742},
  BOOKTITLE = {{PowerTech}},
  ADDRESS = {Kiel, Germany},
  ORGANIZATION = {{IEEE}},
  YEAR = {2025},
  MONTH = Jun,
  HAL_ID = {hal-05262742},
  HAL_VERSION = {v1}
}

@misc{schulman2017proximalpolicyoptimizationalgorithms,
      title={Proximal Policy Optimization Algorithms}, 
      author={John Schulman and Filip Wolski and Prafulla Dhariwal and Alec Radford and Oleg Klimov},
      year={2017},
      eprint={1707.06347},
      archivePrefix={arXiv},
      primaryClass={cs.LG},
      url={https://arxiv.org/abs/1707.06347}
}

@article{xie2025,
  author={Hongbin Xie and Ge Song and Zhuoran Shi and Jingyuan Zhang and Zhenjia Lin and Qing Yu and Hongdi Fu and Xuan Song and Haoran Zhang},
  title={Reinforcement learning for vehicle-to-grid: A review},
  journal={Advances in Applied Energy},
  volume={17},
  year={2025},
  pages={100214},
  doi={10.1016/j.adapen.2025.100214},
  issn={2666-7924},
  url={https://www.sciencedirect.com/science/article/pii/S2666792425000083}
}

@article{DBLP:journals/corr/abs-1209-3352,
  author       = {Shipra Agrawal and
                  Navin Goyal},
  title        = {Thompson Sampling for Contextual Bandits with Linear Payoffs},
  journal      = {CoRR},
  volume       = {abs/1209.3352},
  year         = {2012}
}

@online{statista2025,
  title={Voitures électriques : les modèles qui dominent le marché français en 2025},
  author={Statista},
  year={2025},
  url={https://fr.statista.com/infographie/34343/top-des-ventes-de-voitures-electriques-en-france-selon-le-modele-en-2025/},
  urldate={2025-01-16},
  note={Infographie}
}

@misc{elia2025data,
  title={Photovoltaic Power Generation Data Brussels},
  author={Elia Transmission Belgium SA},
  year={2025},
  note={Elia Open Data License},
  url={https://www.elia.be/en/grid-data/generation-data/solar-pv-power-generation-data}
}

@article{stable-baselines3,
  author  = {Antonin Raffin and Ashley Hill and Adam Gleave and Anssi Kanervisto and Maximilian Ernestus and Noah Dormann},
  title   = {Stable-Baselines3: Reliable Reinforcement Learning Implementations},
  journal = {Journal of Machine Learning Research},
  year    = {2021},
  volume  = {22},
  number  = {268},
  pages   = {1-8},
  url     = {http://jmlr.org/papers/v22/20-1364.html}
}

@misc{caiso2025,
  title={Managing the evolving grid - Renewable curtailment},
  author={{California Independent System Operator}},
  year={2025},
  url={https://www.caiso.com/about/our-business/managing-evolving-grid},
  note={Accessed: October 2025}
}

@misc{bbc2024,
  title={Britain's energy bills problem - and why firms are paid huge sums to stop producing power},
  author={Rowlatt, Justin},
  journal={BBC News},
  year={2024},
  month={June},
  day={9},
  url={https://www.bbc.com/news/articles/cdedjnw8e85o},
  note={October: January 2025}
}

@misc{towers2024gymnasiumstandardinterfacereinforcement,
      title={Gymnasium: A Standard Interface for Reinforcement Learning Environments}, 
      author={Mark Towers and Ariel Kwiatkowski and Jordan Terry and John U. Balis and Gianluca De Cola and Tristan Deleu and Manuel Goulão and Andreas Kallinteris and Markus Krimmel and Arjun KG and Rodrigo Perez-Vicente and Andrea Pierré and Sander Schulhoff and Jun Jet Tai and Hannah Tan and Omar G. Younis},
      year={2024},
      eprint={2407.17032},
      archivePrefix={arXiv},
      primaryClass={cs.LG},
      url={https://arxiv.org/abs/2407.17032}
}

\end{document}